\documentclass[10pt,twocolumn,letterpaper]{article}

\usepackage{cvpr}
\usepackage{times}
\usepackage{epsfig}
\usepackage{graphicx}
\usepackage{amsmath}
\usepackage{amssymb}
\usepackage{multirow}
\usepackage[breaklinks=true,bookmarks=false]{hyperref}

\cvprfinalcopy % *** Uncomment this line for the final submission

\begin{document}

%%%%%%%%% TITLE
\title{Self-Ensembling for 3D Point Cloud Domain Adaptation}

\author{Qing~Li\textsuperscript{1,4}, Xiaojiang~Peng\textsuperscript{1},
Chuan Yan\textsuperscript{2},
Pan Gao\textsuperscript{3},
Qi Hao\textsuperscript{4} \\
\textsuperscript{1}{\small College of Big Data and Internet, Shenzhen Technology University, Shenzhen, China} \\
\textsuperscript{2}{\small Computer Science and Engineering, George Mason University, VA, USA} \\
\textsuperscript{3}{\small Nanjing University of Aeronautics and Astronautics, Nanjing, China} \\
\textsuperscript{4}{\small School of Computer Science and Engineering, Southern University of Science and Technology, Shenzhen, China}\\
% {\small liq36@sustech.edu.cn, pengxiaojiang@sztu.edu.cn, hao.q@sustech.edu.cn}
\and
}

\maketitle
%\thispagestyle{empty}

%%%%%%%%% ABSTRACT
\begin{abstract}
Recently 3D point cloud learning has been a hot topic in computer vision and autonomous driving. Due to the fact that it is difficult to manually annotate a qualitative large-scale 3D point cloud dataset, unsupervised domain adaptation (UDA) is popular in 3D point cloud learning which aims to transfer the learned knowledge from the labeled source domain to the unlabeled target domain. However, the generalization and reconstruction errors caused by domain shift with a simply-learned model are inevitable which substantially hinder the model’s capability from learning good representations. To address these issues, we propose an end-to-end self-ensembling network (SEN) for 3D point cloud domain adaptation tasks.
Generally, our SEN resorts to the advantages of Mean Teacher and semi-supervised learning, and introduces a soft classification loss and a consistency loss, aiming to achieve consistent generalization and accurate reconstruction. In SEN, a student network is kept in a collaborative manner with supervised learning and self-supervised learning, and a teacher network conducts temporal consistency to learn useful representations and ensure the quality of point cloud reconstruction. Extensive experiments on several 3D point cloud UDA benchmarks show that our SEN outperforms the state-of-the-art methods on both classification and segmentation tasks.
Moreover, further analysis demonstrates that our SEN also achieves better reconstruction results.
\end{abstract}

%%%%%%%%% BODY TEXT
\section{Introduction}
Recently, the rapid development of the 3D point cloud has shown crucial importance to many real-world applications in different areas, such as robotics~\cite{pomerleau2015review}, autonomous driving~\cite{chen2017multi}, augmented and virtual reality~\cite{reitinger2007augmented}. The point cloud, as a simple set of points data in a 3D space, is a very convenient format for understanding the 3D world. Compared with 2D data, 3D point clouds can provide detailed geometry position information and easily capture the 3D structure information for the scene, but it is laborious to be processed. The 3D point cloud learning has attracted increasing attention on computer vision areas to process the various vision tasks, such as classification~\cite{qi2017pointnet}, segmentation~\cite{ye20183d}, registration~\cite{lawin2018density} and detection~\cite{vora2020pointpainting}. 
%the various vision tasks, such as classification~\cite{qi2017pointnet, wang2019dynamic}, segmentation~\cite{engelmann2017exploring, ye20183d}, registration~\cite{elbaz20173d, lawin2018density} and detection~\cite{shi2019pointrcnn, vora2020pointpainting}. 

{\bf Related work.} Early studies are mostly based on the hand-crafted features to deal with the 3D point cloud. Driven by the breakthroughs brought by \textit{deep learning}, many studies have actively explored this technique to address these 3D problems on computer vision and robotics tasks. For example, the traditional deep learning is used with the multi-view~\cite{su2015multi} or 3D voxels~\cite{maturana2015voxnet} methods on point cloud tasks. They usually transform 3D points to collections of images or regular 3D voxel before deep model training. However, those pre-processing will cause the loss of geometric information during the data conversion or suffer from high computation or memory consumption.
%They usually transform 3D points to regular 3D voxel grids or  collections of images or  before their training. However, those pre-processing will either cause loss of geometric information during the data conversion or suffered from high computation or memory consumption.The problem with these methods is the lack of geometric information and the high memory consumption. individually computing the feature for each point, thus, it does not consider capturing the local structural information between points.Since PointNet firstly propose that permutation invariance is the fundamental requirement for the 3D point cloud learning task and also theoretically proved the possibility for a neural network to learn on raw 3D points directly.% 
PointNet firstly utilizes deep learning on raw point clouds, which is the bedrock for overcoming the challenge of the irregular nature of point cloud~\cite{qi2017pointnet}. But it is only designed to fuse all features extracted from each point into a global feature, which does not capture the local structural information around a small neighborhood for each point. Subsequently, Qi et al.~\cite{qi2017pointnet++} proposed a hierarchical network PointNet++ to extract local features from the different abstraction levels. Many follow up works such as PointCNN~\cite{li2018pointcnn}, PointConv~\cite{wu2019pointconv} and DGCNN~\cite{wang2019dynamic} also focus on the aggregation of local regions to further improve the feature representation. Despite their impressive success on the point cloud tasks, all of these methods rely on a large amount of labeled data that is time-consuming and expensive to annotate. These make it difficult to keep the model's generalization ability while facing the 3D point clouds in the real world other than the existing dataset. %Although the unlabeled data is relatively easy to obtain abundantly in real applications, it is hard to learn a deep model without annotations for 3D point clouds. 

\begin{figure}
\center
\includegraphics[width=0.9\linewidth]{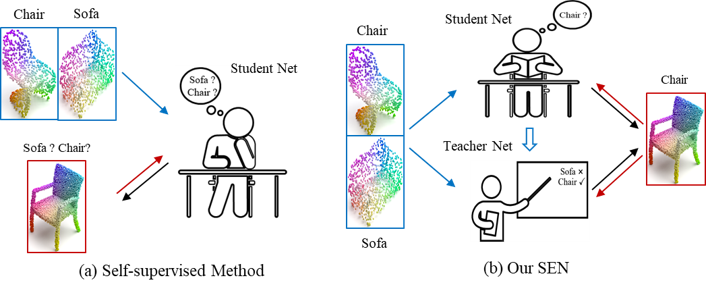}
\caption{Comparison of recent the self-supervised methods and our self-ensembling network (SEN) method  on 3D point cloud UDA tasks.}
\label{fig:motivation}
\end{figure}

The unlabeled data is relatively easy to obtain abundantly in real applications, thus some recent studies attempt \textit{unsupervised learning} for 3D point clouds~\cite{hassani2019unsupervised,chen2020unsupervised}. %It could directly leverage the cheap but informative unlabeled data for training. Therefore, it is necessary and straightforward to explore the strategy of fully unsupervised learning to learn the unlabeled 3D point clouds. Recently, 
To get discriminative information in unsupervised manner, recent works focus on constructing a new structural representation in the form of 3D structure points~\cite{yang2018foldingnet, deng2018ppf}. However, the reconstruction error will eventually affect the quality of feature representation.% Moreover, the reconstructing is applied based on the generated result in previous iterations, which accumulates the reconstruction error during iterations. The problem suggested that it is challenging to learn the discriminative information without the supervision of labeled data.
To address this issue, some other works explore \textit{unsupervised domain adaptation} (UDA) methods on 3D point clouds~\cite{qin2019pointdan, zhao2019multi}, which takes advantage of the labeled source domain dataset to transfer the learned knowledge to the unlabeled target domain dataset. Particularly, there are few research methods based on unsupervised domain adaptation for 3D point cloud classification and segmentation. Currently, PointDAN~\cite{qin2019pointdan} introduces a point-based unsupervised domain adaptation network by aligning the local and global distribution across the different domains for classification tasks. DefRec~\cite{achituve2021self} utilizes the self-supervised tasks to reconstruct deformed input point clouds that produce meaningful representations and obtain a higher classification accuracy than PointDAN on PointDA-10. It suggested that self-supervised learning (SL) is an effective way for the 3D point cloud domain adaptation. However, solely using the SL network often fails to produce promising results with the simple student net, which does not consider the reliability of the network for discriminate feature representations learned from the similar samples on target domain. If we use a teacher net to supervise the student net to give a more distinct reconstruction, we will improve the performance of the whole network. As described in Figure \ref{fig:motivation}, most of the SL methods only rely on a student network to predict the final results. For example, if the chair is similar to the sofa from the source domain, the student net is difficult to discriminate similar samples from the target domain by its ability. Once the teacher net is used to transfer the guidance knowledge to supervise the student net during the process of learning, it can give the accurate judgment of the sample to the chair. 
%The student net's reconstruction of point clouds with similar shapes but different classes without any additional constrain could be ambiguous. For example, the reconstruction of a chair in the target domain might look like both chair and sofa. Therefore, failure to capture the distinct inter-class features will lead to poor classification performance of the student net in the target domain. We introduced the teacher net to guide the student in the target domain so that can help the reconstruction more class-dependent which consequently boosts the student net's classification performance.

%CDN~\cite{su2020adapting} utilizes a domain embedding module to learn a domain vector to characterize the domain attribute information for object detection. SF-UDA3D~\cite{saltori2020sf} is the first source-free unsupervised domain adaptation approach, which is based on pseudo-annotations, reversible scale-transformations, and motion coherency for 3D LiDAR-based detection.%
In this paper, we propose a self-ensembling network (named SEN) for 3D point cloud domain adaptation. Generally, our SEN leveraged the advantages of Mean Teacher and semi-supervised learning to transfer knowledge gained from a labeled source dataset to a different unlabeled target dataset. We introduce a soft classification loss for the source domain learning, which used the teacher model to provide reliable soft labels so that to naturally supervise the student model. We also introduce a consistency loss for target domain learning, which aims to guarantee the predictions to be consistent with the teacher and student network so that to improve the student network's robustness. Moreover, SEN easily bridges the domain gap between supervised and unsupervised representation learning and improves the whole network generalization and prediction by self-ensembling on 3D point clouds.

The main contributions of our paper could be summarized as follows: 1) We propose a novel end-to-end self-ensembling network (SEN) for 3D point cloud domain adaptation tasks. 2) Our SEN resorts to the advantages of Mean Teacher and semi-supervised learning, in which, a student network is kept in a collaborative manner with supervised learning and self-supervised learning, and a teacher network conducts temporal consistency to learn useful representations and improve the accuracy of point clouds reconstruction. 3) We introduce a soft classification loss and a consistency loss to achieve consistent generalization and accurate reconstruction. 4) Our proposed method achieved significant results and gained over the state-of-the-art methods on PointDA-10 and PointSegDA datasets for 3D point cloud domain adaptation tasks.

\section{Approach}
%In this section, we first provided an overview of our proposed method, then present the detailed definitions and notations on 3D point cloud UDA tasks. Finally, we introduced the details of our self-ensembling network (SEN) architecture. It is important to notice that the SEN is mainly described in context of UDA for point clouds classification task, and this method is easy to be extended the UDA for point clouds segmentation task. We pointed out the different loss functions at the end of the section.

\subsection{Overview}
\begin{figure*}
\center
\includegraphics[width=0.75\linewidth]{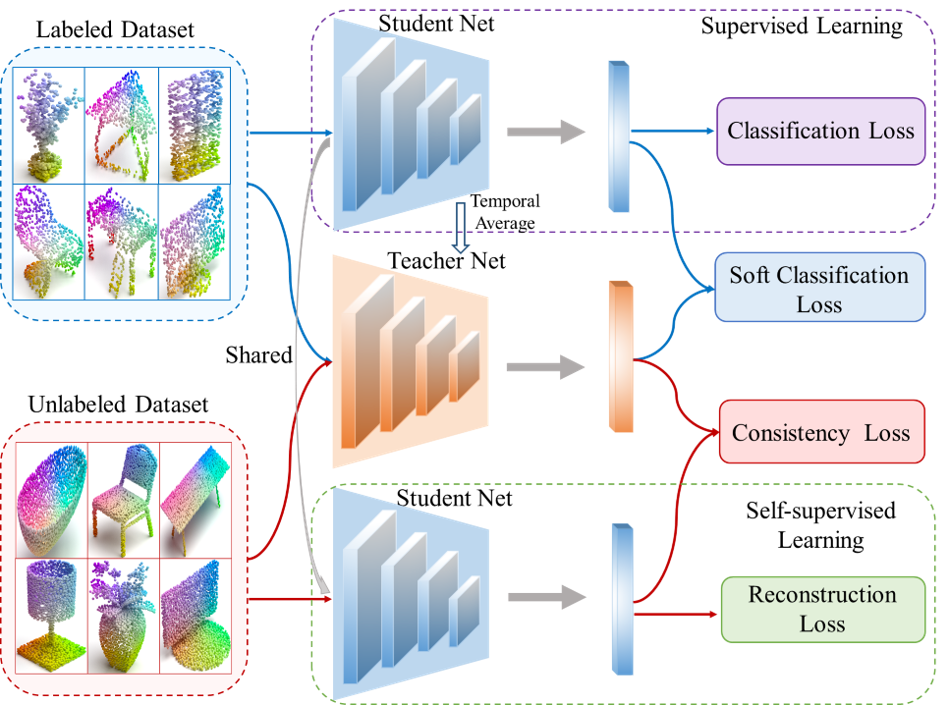}
\caption{The pipeline of our SEN framework for 3D point cloud domain adaptation. It mainly includes two important models (student and teacher net) and joint learning with two branches: 1)the supervised learning on source domain; 2) the self-supervised learning on target domain. Our SEN leverages the teacher and student net to semi-supervised learning by introducing a soft classification loss and a consistency loss, aiming to achieve the consistency generalization and improve the accurate reconstruction.}
\label{fig:framework}
\end{figure*}

% Recent studies suggest that there is acknowledged to adopt the domain adaptation (DA) to alleviate the manual annotation problem in 2D vision tasks. How to use the DA methods to learn the large-scaled or real-world 3D dataset without labels has attracted more and more researchers' attention. There will be more challenges on the 3D domain shift problem compared with the 2D vision tasks.
Recent studies suggest that there are more challenges on the 3D domain shift problem compared with the problem in 2D tasks. To tackle the challenges in unsupervised domain adaptation (UDA) on 3D point clouds, we propose a self-ensembling network (SEN) as illustrated in Figure \ref{fig:framework}. The SEN framework includes two crucial models (student and teacher network) and adopts a jointly training scenario to bridge the gap between supervised learning in the source domains and self-supervised learning in the target domains. The key idea of our method is to leverage the mean-teacher strategy to semi-supervise the whole framework's learning from the source to the target domain. It aims to learn a smoother domain-invariant for improving the consistency of predictions on both domains. In addition, a soft classification loss is introduced to supervise the student net by predicting robust soft labels for the supervised learning part. And a consistency loss is proposed to avoid the error amplification produced from the predictions of student net on the unsupervised learning part. Both of them work together to achieve optimal domain adaptation performances.

\subsection{Definitions and Notations}
About the problem of UDA on 3D point cloud, we give the definitions and notations following~\cite{qin2019pointdan, achituve2021self}. The labelled dataset from the source domain is denoted as $\mathcal{D}_{s}=\left \{ X_{s},Y_{s}\right \} =\left \{ (x_{s}^{i}, y_{s}^{i})\mid _{i}^{N_{s} }\right \}$, where $x_{s}^{i}\in \mathbb{R}^{M\times 3}$, $x_{s}^{i}$ and $ y_{s}^{i}$ represent the $i-th$ training sample point cloud and its associated object label. The $\mathcal{D}_{s}$ has $N_{s}$ sample point clouds with $C_{s}$ unique object class. $M$ is the number of sampling points of one 3D object. We also denote the unlabeled dataset from target domain as $\mathcal{D}_{t}=\left \{ X_{t}\right \} =\left \{ (x_{t}^{i})\mid _{i}^{N_{t} }\right \}$, where $x_{t}^{i}\in \mathbb{R}^{M\times 3}$, and $x_{t}^{i}$ is the $i-th$ training sample point cloud and without any ground-truth object label, $ N_{t}$ is the number of sample point cloud on target domains. Followed the previous methods, the target domain task is assumed to be the same with the source domain task. It means that the source label space $Y_{s}$ is shared with the target label space $Y_{t}$. Our ultimate goal is to leverage the labelled sample point clouds in $\mathcal{D}_{s}$ and the unlabeled sample point clouds in $\mathcal{D}_{t}$ to jointly learn a transferred model for the target-domain $\mathcal{D}_{t}$.

\subsection{Self-ensembling Network Architecture }

\begin{figure}
\center
\includegraphics[width=0.9\linewidth]{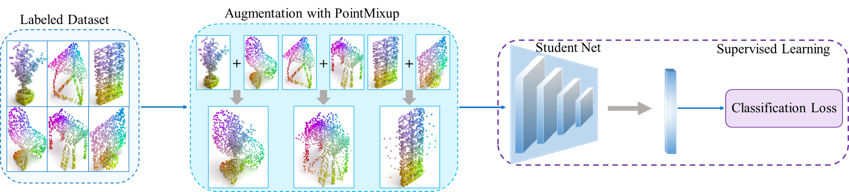}
\caption{The branch of supervised learning on source domain. It uses an optional augmentation scheme to increase the numbers and diversity of source domain examples so that to improve the network generalization ability.}
\label{fig:supervised}
\end{figure}

{\bf Supervised learning for source domain.} The goal of domain adaptive on 3D point cloud is to learn a discriminative feature extractor $f(\cdot |\theta )$ for $\mathcal{D}_{t}$. The student network is first trained on the source domain, where the labels of source objects are available. Therefore, the training on $\mathcal{D}_{s}$ can consider as a standard point cloud classification problem. We optimize the network by using the cross-entropy (CE) loss, which is formulated as follows:

\begin{equation}\label{eq:positive}
\mathcal{L}_{s} = -\frac{1}{N_{s}} \sum_{i=1}^{N_{s}} \log{p(y_{s}^{i}\mid x_{s}^{i})}
\end{equation}
where $ p(y_{s}^{i}\mid x_{s}^{i})$ is the predicted probability that the source object $x_{s}^{i}$ belongs to class $ y_{s}^{i}$. The branch on supervised learning is aptly described in Figure \ref{fig:supervised}. The supervised learning on labeled source data ensures the performance of $f(\cdot |\theta )$ on the same distributed valuation and testing set. However, the performance will decrease when we test on the different distribution target data. We will introduce a the self-ensembling method to overcome this problem by considering the domain consistency variations of both domains.

{\bf PointMixup for point cloud augmentation.} Data augmentation is a common technique to improve the model generalization by artificially increasing the similar but different training samples. It is appropriate to alleviate the limitation of the small size and the variance of the original training data.
%  Mixup~\cite{zhang2018mixup} is a simple and effective data augmentation method, has been feasibly applied to various 2D vision tasks. such as image classification~\cite{he2019bag} object detection~\cite{zhang2019bag}and segmentation~\cite{french2020semi}.Additionally, the mixup also extends to address the UDA problem on 2D or 3D representation learning tasks~\cite{wu2020dual, yan2020improve, achituve2021self}. The process of the mixup training can help the model infers consistent predictions in the latent space and directly improve the generalization of the classifier when applied to the target domain.
Due to the 3D point cloud is limited by the number of training samples and an immense augmentation space, Chen et al.~\cite{chen2020pointmixup} proposed to use the simple PointMixup (PM) as an interpolation method to generates new examples for points clouds classification tasks. Mixup is a feasible augmentation method by adding the diversity of training samples to reduce the generalization error.  % Especially, let $(x_{i},y_{i})$ and $(x_{j},y_{j})$ denote as a pair of sample-label tuples, we chose a number of points from each of the pair tuples to generate a new mixup-sample and its label as followed:%
In this paper, we utilize the augmentation method of PM to assist our SEN framework in learning more 3D representation for point clouds UDA tasks. Let $(x_{i}, y_{i})$ and $(x_{j},y_{j})$ denote as a pair of sample-label tuples, we use the point sampled interpolation method to generate a new sampled set and its labels as: 

\begin{equation}\label{eq:mixup}
\begin{array}{l}
x^{'}  = FPS((1-\gamma ) \cdot M, x_{i}) \cup FPS(\gamma \cdot M , x_{j})\\
y^{'}  = (1-\gamma )\cdot y_{i} \cup \gamma y_{j}
\end{array}
\end{equation}
where $FPS((1-\gamma )N_{i}, x_{i})$ computes a new sampled subset by using farthest point sampling (FPS) to sample from $x_{i}$ with the numbers of $(1-\gamma ) \cdot M$ points, $FPS(\gamma \cdot M, x_{j})$ is the same sampled from $x_{j}$ with the numbers of $\gamma \cdot M$ points. The $M$ has defined as the number of sampling points about the sample $x_{i}$. The $\gamma\in \left [ 0,1 \right ]$ is the interpolation weights coefficient, and $ \gamma\sim Beta(\alpha,\alpha )$, in which $\alpha $ is set as 0.2. Based on PM method, our SEN use this augmentation strategy in the source domain for adding the diversity of training samples. We chose the same cross-entropy loss function to computer the loss, and define as $\mathcal{L}_{s}^{'}$.%to add into the original $ \mathcal{L}_{s}$

% \begin{equation}\label{eq:mixup}
% \begin{array}{l}
% x^{'}  = \hat{x}_{i} \cup \hat{x}_{j} \\
% y^{'}  = (1-\gamma )y_{i} \cup \gamma y_{j}
% \end{array}
% \end{equation}
% where $\hat{x}_{i}$ is calculated using farthest point sampling (FPS) of the input point cloud from $x_{i}$ with the number of $(1-\gamma )\cdot M$ points, $\hat{x}_{j})$ is computed using FPS of the input point cloud from $x_{j}$ with the number of $\gamma \cdot  M$ points. The $M$ has defined as the number of sampling points about the sample $x_{i}$. The $\gamma\in \left [ 0,1 \right ]$ is the interpolation weighting coefficient, and $ \gamma\sim Beta(\alpha,\alpha )$, in which $\alpha $ keeps to set as 0.2. Based on Point cloud Mixup, our SEN use this augmentation strategy in the source domain for adding the diversity of training samples. We chose the same cross-entropy loss function to computer the loss, and define as $ \mathcal{L}_{s}^{'} $ to instead of the original $ \mathcal{L}_{s}$.

\begin{figure}
\center
\includegraphics[width=0.9\linewidth]{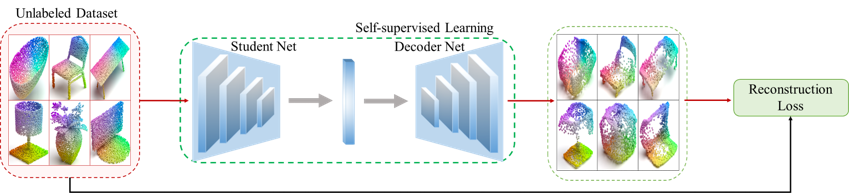}
\caption{The branch of self-supervised learning on the target domain. It is trained like a simple auto-encoder, which only contains a student net and a decoder net.}
\label{fig:self-supervised}
\end{figure}

{\bf Unsupervised learning for target domain.} Due to the training samples on target domain $\mathcal{D}_{t}$ without labels, we employ a simple unsupervised method, named as self-supervised learning, which is used the consistent structure points as structural representation directly on 3D point clouds. Recently, some works has been proposed to generate accurate reconstruction of 3D shapes for unsupervised learning of point clouds. We choose the widely-used encoder-decoder architecture (like FoldingNet~\cite{yang2018foldingnet}) for the branch of self-supervised learning, which is shown in Figure \ref{fig:self-supervised}. The encoder adopts the same student network with shared model parameters in the source domain. Thus, we use the feature extractor $f(\cdot |\theta )=f(x_{i}^{t} |\theta)$ to obtain the feature representation $f_{i}^{t}=f(x_{i}^{t} |\theta)$ in the target domain. And the decoder is inspired by the FoldingNet decoder architecture. It consists a concatenation of 4-layer multi-layer perceptrons (MLP) to reconstruct the original point set as the $\mathcal{Z }_{t}=\left \{ \hat{X}_{t}\right \}$ and the point $\hat{x}_{t}^{i}\in \mathbb{R}^{N\times 3}$.

We propose a reconstruction loss to train the network in an unsupervised manner, which is used to penalize the difference between the reconstructed and the original 3D point cloud. It computes the Chamfer distance (CD)~\cite{fan2017point} to constrain the predicted structure points $\hat{X}_{t}$ to be close to the input points $X_{t}$. Hence, we optimize the reconstruction loss as follows:

\begin{equation}\label{eq:chamfer}
% \begin{array}{l}
\mathcal{L}_{t}(\hat{X}_{t}, X_{t})= \sum_{\hat{x}_{t}^{i}\in \hat{X}_{t}}\min_{x_{t}^{j}\in X_{t}}\left \|\hat{x}_{t}^{i}- x_{t}^{j} \right \| _{2}^{2} + \sum_{x_{t}^{j}\in X_{t}}\min_{x_{t}^{i}\in \hat{X}_{t}}\left \|\hat{x}_{t}^{i}- x_{t}^{j} \right \| _{2}^{2}
% \end{array}
\end{equation}

{\bf Self-ensembling with Mean-Teacher method.} The mean-teacher method has been extended to address the UDA problem by minimizing the discrepancy between predictions on the source and target domain. It also consists of two models with identical architecture, a student network, and a teacher network. The main idea of mean-teacher is to "smooth" the student network decision boundary between similar samples by the average of previous predictions from the teacher network. In order to constrain the student network to have consistent predictions with the teacher network, we propose the SEN framework to leverage the mean-teacher network for better performance on 3D point clouds UDA tasks. In SEN, the student is trained with the collaboration of supervised learning and self-supervised learning. The teacher network aims to supervise the training of the student network by predicting robust soft labels on the source domain and by calculating the feature consistency on the target domain. Therefore, the SEN framework not only bridges the gap between supervised and unsupervised representation learning but also improves the consistency of generalization and the accurate reconstruction. To avoid the two networks (teacher-student) collaboratively bias each other, we introduce soft classification loss for the branch of supervised learning and a consistency loss for the branch of unsupervised learning.

During in the SEN framework, the student network is trained using gradient descent, while the teacher network is an average of consecutive student network. Therefore, we update the teacher’s weights ${\theta}'$ as an exponential moving average (EMA) of the student’s weights $\theta$ to ensemble the information in different training steps~\cite{tarvainen2017mean}. Thus, the temporally average model of the student network at training step T are calculated as:

\begin{equation}\label{eq:mt}
{\theta}'_{t}= \alpha {\theta}'_{t-1}+(1-\alpha)\theta_{t} 
\end{equation}
where $ {\theta}'_{t-1}$ indicate the temporal average parameters of the teacher model at training step $t-1$, $\alpha$ is the ensembling momentum hyperparameter to be within the range [0, 1). During training on supervised learning, the teacher network leverages the generated soft pseudo label for supervising the student network. We denote the soft classification loss function on the source domain, as

\begin{equation}\label{eq:soft}
\mathcal{L}_{soft}= -\frac{1}{N_{s}} \sum_{i=1}^{N_{s}} p(x_{s}^{i}\mid {\theta}') \log{p(x_{s}^{i}\mid \theta)} 
\end{equation}

The student network shared the same parameters for training on the source and target domain, the teacher network is also used to generate supervisions for both domains and therefore avoid generalization and reconstruction error caused by domain shift. In order to further mitigate the effects caused by domain shift, we propose a consistency loss to maintain the student network's instance discrimination ability on the target domain. More formally, the consistency loss is computed by the distance between the feature $f_{i}^{t}=f(x_{i}^{t} |\theta)$ of the student network and the feature $f_{i}^{t}=f(x_{i}^{t} | {\theta}')$ of the teacher network as follows:

\begin{equation}\label{eq:cons}
\mathcal{L}_{cons}= \left \| f(x_{i}^{t} | \theta)-f(x_{i}^{t} | {\theta}')\right \| _{2}^{2}  
\end{equation}
where $\left \| \cdot \right \| $ denotes the $L_{2}-$norm distance, $f(x_{i}^{t} |\theta)$ has been denoted as the feature representation in the target domain, which is obtained by the feature extractor $f(\cdot |\theta )$,the similar on $ f(x_{i}^{t} |{\theta}')$ is extracted by $f(\cdot |{\theta}' )$.

{\bf Overall loss.} The proposed SEN framework is jointly trained in the source and target domain data by optimizing all losses with an end-to-end manner. Finally, the overall loss on classification can be written as:

\begin{equation}\label{eq: classification }
\mathcal{L}_{SEN}^{c}=\lambda(\mathcal{L}_{s} + \mathcal{L}_{soft})+ \mathcal{L}_{t}+\mathcal{L}_{cons}
\end{equation}

When we focus on the loss design for segmentation task, we don't need to compute the soft classification loss at the source domain any more but replace with the segmentation consistency on the source domain. Therefore, we use the same consistency loss instead of the soft classification loss in the source domain. If we use the PM for point cloud augmentation, we will use the new data as the source data to compute the CE loss of $\mathcal{L}_{s}^{'}$ to instead of $\mathcal{L}_{s}$ in our SEN framework.

\section{Experiments}
\subsection{Datasets and implementation detail}
We study UDA learning and evaluate our proposed SEN on 3D point clouds classification and segmentation tasks. Each of them includes a set of DA datasets, PointDA-10~\cite{qin2019pointdan} and PointSegDA~\cite{achituve2021self}, respectively.  

\begin{figure}
\center
\includegraphics[width=0.75\linewidth]{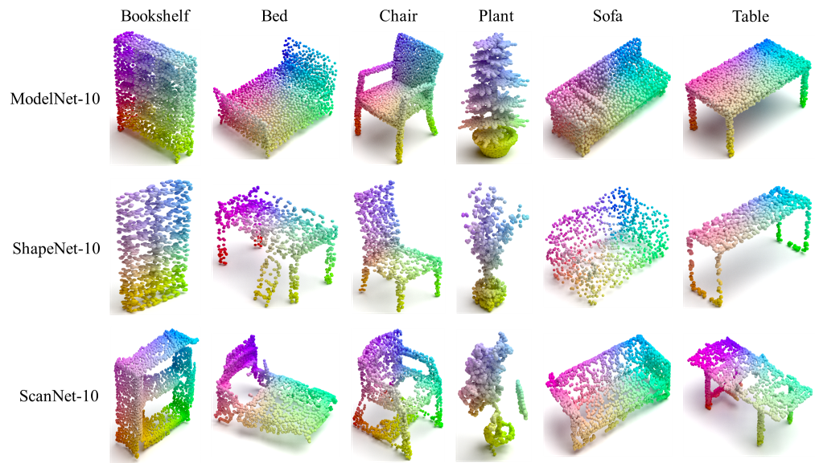}
\caption{A comparison of examples from three subset datasets: ModelNet-10, ShapeNet-10, and ScanNet-10}
\label{fig:cl_data}
\end{figure}

{\bf PointDA-10 datase.} PointDA-10 includes three subsets with different characteristics for domain adaptation. They are named as ModelNet-10, ShapeNet-10, and ScanNet-10. All of these subsets include the same 10 shared classes (i.e., bookshelf, bed, chair, table). Figure \ref{fig:cl_data} presents some sample visualizations for a comparison of typical shapes from the three subsets. {\bf ModelNet-10} contains 5039 meshed CAD models in which 4183 models are used for training and 856 models for testing. {\bf ShapeNet-10} has 19,870 unique 3D models, including 17,378 samples for training and 2492 samples for testing. {\bf ScanNet-10} contains 7869 objects and splits into 6110 train objects and 1769 test objects.
%{\bf PointDA-10 datase.} PointDA-10 includes three subsets with different characteristics for domain adaptation. They are selected from three widely adopted 3D benchmark datasets and are named as ModelNet~\cite{wu20153d}, ShapeNet~\cite{chang2015shapenet}, and ScanNet~\cite{dai2017scannet}. All of these subsets include the same 10 shared classes (i.e., bookshelf, bed, chair, table). Figure \ref{fig:cl_data} presents some sample visualization for a comparison of typical shapes from the three datasets.
% {\bf ModelNet-10} is extracted from the ModelNet40 with the clean 3D CAD models. And it contains 5039 meshed CAD models from 10 different object categories out of which 4183 models are used for training and 856 models for testing. 
% {\bf ShapeNet-10} is selected from the ShapeNetCore, which is a subset of the full ShapeNet dataset with single clean 3D models and has manually verified categories. It covers the same 10 common object categories with about 19,870 unique 3D models, including 17,378 samples for training and 2492 samples for testing.
% {\bf ScanNet-10} is isolated from the original ScanNet, which is a large-scale and real-world indoor scene. The dataset provides ground truth labels for 3D objects classification. However, many objects have missing parts and are sampled sparsely. It contains 7869 objects and splits into 6110 train objects and 1769 test objects.

\begin{figure}
\center
\includegraphics[width=0.7\linewidth]{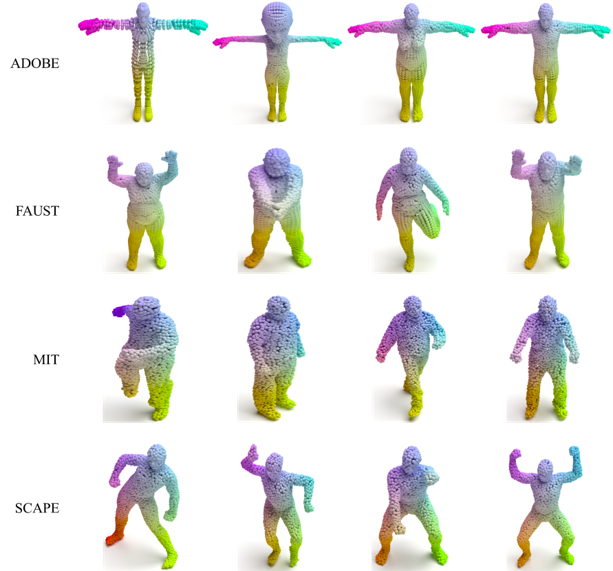}
\caption{A comparison of typical point clouds from four subsets: ADOBE, FAUST, MIT, and SCAPE}
\label{fig:sg_data}
\end{figure}

{\bf PointSegDA dataset.} PointSegDA consists of four subsets, which are ADOBE, FAUST, MIT, and SCAPE, respectively. These subsets are mainly about the meshes of human models for body parts segmentation tasks. They contain 8 shared classes (like feet, hand, head, etc.), but they have different body pose, shape, and point distribution. We visualize a comparison of typical body poses from all the datasets mentioned above, seen in Figure \ref{fig:sg_data}. {\bf ADOBE} dataset includes 42 samples, which has been split to 29 for training, 8 for testing, and the rest for validation. {\bf FAUST} dataset contains 100 objects (70 train, 20 test, 10 validation). {\bf MIT} dataset has 169 samples in the sets, with 118 train, 34 test, and 17 validations. The subset of {\bf SCAPE} has a total of 71 bodies, including 50 trains, 14 tests, and 7 validations.

\begin{table*}[htp]
\centering
\caption{Performance ($\%$) comparison of our framework with state-of-the-art methods on PointDA-10 dataset.}\label{tab:table_c}
\footnotesize
\scalebox{0.9}{
\begin{tabular}{l|l|l|l|l|l|l} 
\hline
 \multicolumn{1}{c|}{\multirow{2}{*}{Method}} &\textbf{ModelNet to}& \textbf{ModelNet to} & \textbf{ShapeNet to} & \textbf{ShapeNet to} & \textbf{ScanNet to}& \textbf{ScanNet to} \\
 \multicolumn{1}{c|}{} &\textbf{ShapeNet} & \textbf{ScanNet} & \textbf{ModelNet} & \textbf{ScanNet} & \textbf{ModelNet} & \textbf{ShapeNet} \\
\hline
\hline
 Supervised-T & $93.9~\pm~0.2$ & $78.4~\pm~0.6$ & $96.2~\pm~0.1$ & $78.4~\pm~0.6 $& $96.2~\pm~0.1$& $93.9~\pm~0.2$ \\
 Supervised & $89.2~\pm~0.6$ & $76.2~\pm~0.6$ &$ 93.4~\pm~0.6$ & $74.7~\pm~0.7$& $93.2~\pm~0.3$& $88.1~\pm~0.7$\\
 \hline
 Unsupervised & $83.2~\pm~0.7$ & $43.8~\pm~2.3$ &$ 75.5~\pm~1.8$ & $42.5~\pm~1.4$& $63.8~\pm~3.9$& $64.2~\pm~0.8$\\
 DANN & $75.3~\pm~0.6$ & $41.5~\pm~0.2$ &$ 62.5~\pm~1.4$ & $46.1~\pm~2.8$& $53.3~\pm~1.2$& $63.2~\pm~1.2$\\
 PointDANN & $82.5~\pm~0.8$ & $47.7~\pm~1.0$ &$ 77.0 ~\pm~0.3$ & $48.5~\pm~2.1$& $55.6~\pm~0.6$& $67.2~\pm~2.7$\\
 RS & $81.5~\pm~1.2$ & $35.2~\pm~5.9$ &$ 71.9~\pm~1.4$ & $39.8~\pm~0.7$& $61.0~\pm~3.3$& $63.6~\pm~3.4$\\
 DAE-Global & $\textbf{83.5}~\pm~\textbf{0.8}$ & $42.6~\pm~1.4$ &$ 74.8~\pm~0.8$ & $45.5~\pm~1.6$& $64.9~\pm~4.4$& $67.3~\pm~0.6$\\
 DAE-Point & $82.5~\pm~0.4$ & $40.2~\pm~1.6$ &$ 76.4~\pm~0.7$ & $50.2~\pm~0.5$& $66.3~\pm~1.5$& $66.1~\pm~0.5$\\
 DefRec & $83.3~\pm~0.2$ & $46.6~\pm~2.0$ &$ \textbf{79.8}~\pm~\textbf{0.5}$ & $49.9~\pm~1.8$& $70.7~\pm~1.4$& $64.4~\pm~1.2$\\
 DefRec~+~PCM & $81.7~\pm~0.6$ & $51.8~\pm~0.3$ &$ 78.6~\pm~0.7$ & $\textbf{54.5}~\pm~\textbf{0.3}$& $73.7~\pm~1.6$& $71.4~\pm~1.4$\\
 \hline
  SEN & $82.6~\pm~0.1$ & $50.1~\pm~1.9$ & $75.9~\pm~0.5$ & $50.7~\pm~0.4$ & $74.2~\pm~0.3$ & $69.1~\pm~0.1$ \\
 SEN~+~PM & $83.3~\pm~0.3$ & $\textbf{53.1}~\pm~\textbf{2.5}$ & $70.3~\pm~1.0$ & $54.5~\pm~0.2$ & $\textbf{74.5}~\pm~\textbf{1.5}$ & $\textbf{72.8}~\pm~\textbf{0.6}$ \\
 \hline
\end{tabular}}
\end{table*}

{\bf Implementation details.} We implement our method with Pytorch. During the training, we augment the point clouds on-the-fly with random jitter for each point using standard deviation and clip parameters of 0.01 and 0.02 respectively. We adopt the DGCNN as the base network architecture for a fair comparison.  For network optimization, we use the Adam optimizer with a cosine annealing learning rate scheduler. All models adopted an NVIDIA TITAN GPU for training. In a classification task, the training batch size of both domains, the total epochs numbers, and the initial learning rate are set as 32, 150, and $10^{-3}$, respectively. The parameter of weight $\lambda$ is set to the default value of 0.2. About the segmentation task, the training batch size for both domain, the total epochs numbers and the initial learning rate is fixed to 16, 200 and $10^{-3}$, respectively. The parameter of weight $\lambda$ is selected with the default value of 0.05.

\begin{table*}[htp]
\centering
\footnotesize
\caption{Performance ($\%$) comparison of our framework with state-of-the-art methods on PointSegDA dataset.}\label{tab:table_s}
\scalebox{0.68}{
\begin{tabular}{c|l|l|l|l|l|l|l|l|l|l|l|l|l} 
\hline
% \hline%{\textbf{Method}}%
 \multicolumn{1}{c|}{\multirow{2}{*}{Method}} &\textbf{FAUST}&\textbf{FAUST}&\textbf{FAUST}&\textbf{MIT}&\textbf{MIT}&\textbf{MIT}&\textbf{ADOBE}&\textbf{ADOBE}&\textbf{ADOBE}&\textbf{SCAPE}&\textbf{SCAPE}&\textbf{SCAPE}&\textbf{} \\
  \multicolumn{1}{c|}{\textbf{}} &\textbf{to}&\textbf{to}&\textbf{to}&\textbf{to}&\textbf{to}&\textbf{to}&\textbf{to}&\textbf{to}&\textbf{to}&\textbf{to}&\textbf{to}&\textbf{to}&\textbf{Avg.} \\
  \multicolumn{1}{c|}{} &\textbf{ADOBE}&\textbf{MIT}&\textbf{SCAPE}&\textbf{ADOBE}&\textbf{FAUST}&\textbf{SCAPE}&\textbf{FAUST}&\textbf{MIT}&\textbf{SCAPE}&\textbf{ADOBE}&\textbf{FAUST}&\textbf{MIT}& \textbf{} \\
\hline
\hline
Supervised-T & $84.0~\pm~1.8$ & $84.0~\pm~1.8$ & $84.0~\pm~1.8$ & $81.8~\pm~0.3$& $81.8~\pm~0.3$& $81.8~\pm~0.3$& $80.9~\pm~7.2$& $80.9~\pm~7.2$& $80.9~\pm~7.2$& $82.4~\pm~1.2$& $82.4~\pm~1.2$& $82.4~\pm~1.2$& $82.3~\pm~2.6$\\
Unsupervised& $78.5~\pm~0.4$ & $60.9~\pm~0.6$ & $66.5~\pm~0.6$ & $26.6~\pm~3.5$& $33.6~\pm~1.3$& $69.9~\pm~1.2$& $38.5~\pm~2.2$& $31.2~\pm~1.4$& $30.0~\pm~3.6$& $\textbf{74.1}~\pm~\textbf{1.0}$& $68.4~\pm~2.4$& $64.5~\pm~0.5$& $53.6~\pm~1.6$\\
Adapt-SegMap & $70.5~\pm~3.4$ & $60.1~\pm~0.6$ & $65.3~\pm~1.3$ & $49.1~\pm~9.7$& $54.0~\pm~0.5$& $62.8~\pm~7.6$& $44.2~\pm~1.7$& $35.4~\pm~0.3$& $35.1~\pm~1.4$& $70.1~\pm~2.5$& $67.7~\pm~1.4$& $63.8~\pm~1.2$& $56.5~\pm~2.6$\\
RS & $78.7~\pm~0.5$ & $60.7~\pm~0.4$ & $66.9~\pm~0.4$ & $59.6~\pm~5.0$& $38.4~\pm~2.1$& $70.4~\pm~1.0$& $44.0~\pm~0.6$& $30.4~\pm~0.5$& $36.6~\pm~0.8$& $70.7~\pm~0.8$& $\textbf{73.0}~\pm~\textbf{1.5}$& $65.3~\pm~0.1$& $57.9~\pm~1.1$\\
DefRec & $\textbf{79.7}~\pm~\textbf{0.3}$ & $61.8~\pm~0.1$ & $\textbf{67.4}~\pm~\textbf{1.0}$ & $67.1~\pm~1.0$& $40.1~\pm~1.4$& $72.6~\pm~0.5$& $42.5~\pm~0.3$& $28.9~\pm~1.5$& $32.2~\pm~1.2$& $66.4~\pm~0.9$& $72.2~\pm~1.2$& $66.2~\pm~0.9$& $58.1~\pm~0.9$\\
DefRec~+~PCM & $78.8~\pm~0.2$ & $60.9~\pm~0.8$ & $63.6~\pm~0.1$ & $48.1~\pm~0.4$& $48.6~\pm~2.4$& $70.1~\pm~0.8$& $46.9~\pm~1.0$& $33.2~\pm~0.3$& $37.6~\pm~0.1$& $66.3~\pm~1.7$& $66.5~\pm~1.0$& $62.6~\pm~0.2$& $56.9~\pm~0.7$\\
\hline
SEN & $\textbf{79.7}~\pm~\textbf{0.3}$ & $\textbf{63.9}~\pm~\textbf{0.4}$ & $65.1~\pm~0.6$ & $64.7~\pm~1.7$ & $\textbf{58.7}~\pm~\textbf{0.1}$ & $\textbf{76.2}~\pm~\textbf{0.3}$ & $40.5~\pm~0.1$ & $32.0~\pm~0.6$ & $31.9~\pm~0.6$ & $64.9~\pm~0.7$ & $61.0~\pm~1.1$ & $\textbf{72.2}~\pm~\textbf{0.1}$ & $59.2~\pm~0.5$ \\
SEN~+~PM & $75.9~\pm~0.9$ & $63.1~\pm~0.4$ & $62.8~\pm~0.3$ & $\textbf{78.7}~\pm~\textbf{0.5}$ & $56.4~\pm~1.0$ & $75.7~\pm~0.1$ & $\textbf{51.8}~\pm~\textbf{0.8}$ & $\textbf{43.1}~\pm~\textbf{0.4}$ & $\textbf{43.2}~\pm~\textbf{1.0}$ & $64.7~\pm~0.3$ & $67.1~\pm~0.1$ & $69.5~\pm~0.1$ & $\textbf{62.7}~\pm~\textbf{0.5}$ \\
\hline
\end{tabular}}
\end{table*}

{\bf Data processing.} Following the previous studies~\cite{wang2019dynamic, achituve2021self}, we assume that all datasets are under an ideal condition where point clouds are uniformly sampled along the accordant direction as the pre-aligned setting for training and evaluation. In the classification task, especially for the ModelNet, where the object instances have any arbitrary rotation only along the Z-axis. We kept the same direction in ShapeNet and ScanNet by rotating 90 degrees around their X-axis. We also used 1024 points for all these subsets by the farthest point sampling method as described in~\cite{qi2017pointnet}. Focus on the segmentation task, we sampled 2048 points by the farthest point sampling and rotate all shapes from different subsets to the same direction by the positive Z-axis.

{\bf Evaluation metrics.} We followed the standard training/ validation /test split and evaluated the test evaluation settings. For classification evaluation metrics, we use the mean accuracy and standard error of the mean (SEM) across repeated three times of testing with different seeds. For segmentation evaluation metrics, we report the mean IoU (Intersection-over-Union) by averaged over three times of testing.

\subsection{Comparison with state of the arts on classification}
To investigate the effect of the collaborative learning based on our proposed SEN, we first compare SEN with three baselines: 1) the basic supervised learning on the target dataset; 2) direct training with the supervised learning on both source and target dataset; 3) supervised learning on the source domain and directly tested on the target domain. Then, we compare our approach with the state-of-the-art UDA methods for 3D point cloud including: \textit{DANN}~\cite{ganin2016domain}, \textit{PointDAN}~\cite{qin2019pointdan}, \textit{RS}~\cite{sauder2019self}, \textit{(DAE)-Global}~\cite{hassani2019unsupervised} and \textit{DefRec}~\cite{achituve2021self}. The experiment results on the three subsets of domain adaptation tasks are shown in Table \ref{tab:table_c}

As can be seen, the \textit{supervised-T}, in the first row, represents the upper bound performance of the target domain dataset. The row of second and third represent the performance of domain diversity and domain-shift between the source and target domain, respectively. We find that directly transfer the learned knowledge from the source domain to the target domain is difficult due to the clear diversity between them. Compared with our SEN, the third base line model only used the pertained model on the source domain, which doesn't mine the discriminative information in the target domains, so that it demonstrates the effectiveness of our SEN learning in both domains. Compared with the state-of-the-art UDA methods, our proposed method outperforms these methods with a significant margin on three out of six domain gap scenarios. Meanwhile, the DAE-Global obtains the highest accuracy on ModelNet to ShapeNet and DefRec achieves the highest accuracy on ShapeNet to ModelNet. Compared with the DefRec (as a representative of the pure self-supervised method), our SEN uses the self-ensembling to conduct the temporal consistency to learn the useful feature representations and ensure the quality of point clouds reconstruction. Moreover, we also adopt the PM (PointMixup) models to explore the latent representations on 3D point clouds to improve the performance of our SEN framework.

\begin{table*}
\centering
\caption{Performance ($\%$) comparison of our framework with additional models on PointDA-10 dataset.}\label{tab:table_ca}
\footnotesize
\scalebox{0.8}{
\begin{tabular}{l|l|l|l|l|l|l|l} 
\hline
 \multicolumn{1}{c|}{\multirow{2}{*}{Method}} &\textbf{ModelNet to}& \textbf{ModelNet to} & \textbf{ShapeNet to} & \textbf{ShapeNet to} & \textbf{ScanNet to}& \textbf{ScanNet to}& \multirow{2}{*}{\textbf{Avg.}} \\
 \multicolumn{1}{c|}{} &\textbf{ShapeNet} & \textbf{ScanNet} & \textbf{ModelNet} & \textbf{ScanNet} & \textbf{ModelNet} & \textbf{ShapeNet} & \textbf{} \\
\hline
\hline
 DefRec & $83.3~\pm~0.2$ & $46.6~\pm~2.0$ &$ \textbf{79.8}~\pm~\textbf{0.5}$ & $49.9~\pm~1.8$& $70.7~\pm~1.4$& $64.4~\pm~1.2$ & $65.8~\pm~1.2$\\
 DefRec+PCM  & $81.7~\pm~0.6$ & $51.8~\pm~0.3$ &$ 78.6~\pm~0.7$ & $54.5~\pm~0.3$& $73.7~\pm~1.6$& $71.4~\pm~1.4$ & $68.6~\pm~0.8$\\
 DefRecS/T+PCM &  $82.6~\pm~0.6$ & $53.1~\pm~1.0$ &$ 78.3~\pm~1.0$ & $51.5~\pm~0.9$& $72.0~\pm~0.5$& $74.4~\pm~0.8$ & $68.7~\pm~0.8$\\
 \hline
 SEN+DefRec & $82.4~\pm~0.8$ & $51.5~\pm~1.3$ & $75.7~\pm~0.6$ & $49.2~\pm~0.5$ & $72.2~\pm~1.3$ & $67.3~\pm~0.5$ & $66.4~\pm~0.8$ \\
 SEN+DefRec~+~PM & $83.0~\pm~0.2$ & $53.8~\pm~1.1$ & $72.2~\pm~0.2$ & $\textbf{55.0}~\pm~\textbf{0.2}$ & $75.0~\pm~0.8$ & $75.6~\pm~0.6$ & $69.1~\pm~0.5$ \\
 SEN+DefRec S/T+PM & $ \textbf{83.7}~\pm~\textbf{0.1}$ & $ \textbf{54.0}~\pm~\textbf{0.8}$ & $73.6~\pm~1.4$ & $54.7~\pm~0.5$ & $\textbf{77.1}~\pm~\textbf{0.5}$ & $\textbf{76.3}~\pm~\textbf{0.7}$ & $\textbf{69.9}~\pm~\textbf{0.7}$ \\
 SEN & $82.6~\pm~0.1$ & $50.1~\pm~1.9$ & $75.9~\pm0.5$ & $50.7~\pm~0.4$ & $74.2~\pm~0.3$ & $69.1~\pm~0.1$ & $67.1~\pm~0.5$ \\
 SEN~+~PM & $83.3~\pm~0.3$ & $53.1~\pm~2.5$ & $70.3~\pm~1.0$ & $54.5~\pm~0.2$ & $74.5~\pm~1.5$ & $72.8~\pm~0.6$ & $68.1~\pm~1.0$ \\
 SEN+PM+Rec S/T & $83.5~\pm~0.9$ & $52.9~\pm~1.0$ & $71.2~\pm~2.2$ & $54.0~\pm~0.5$ & $74.6~\pm~0.8$ & $72.5~\pm~0.8$ & $68.1~\pm~1.0$ \\
 \hline
\end{tabular}}
\end{table*}

\subsection{Comparison with state of the arts on segmentation}
We also conduct the comparison with the state-of-the-art 3D point cloud segmentation methods on PointSegDA, which is a domain adaptation dataset on segmentation tasks. Due to a limited number of works report performance on PointSegDA, we compared our SEN with the baseline (supervised and unsupervised method) and RS~\cite{sauder2019self}, Adapt-SegMap~\cite{tsai2018learning}, DefRec~\cite{achituve2021self}. As shown in Table \ref{tab:table_s}, our approach outperforms most existing methods by large margins under both unsupervised and domain adaptation methods. The PM models help boost the performance on most cross-domain segmentation tasks, especially for MIT to ADOBE and the average results.

\begin{table*}[htp]
\centering
\footnotesize
\caption{Performance ($\%$) comparison of our framework with additional models on PointSegDA dataset.}\label{tab:table_sa}
\scalebox{0.68}{
\begin{tabular}{c|l|l|l|l|l|l|l|l|l|l|l|l|l} 
\hline
% \hline%{\textbf{Method}}%
 \multicolumn{1}{c|}{\multirow{2}{*}{Method}} &\textbf{FAUST}&\textbf{FAUST}&\textbf{FAUST}&\textbf{MIT}&\textbf{MIT}&\textbf{MIT}&\textbf{ADOBE}&\textbf{ADOBE}&\textbf{ADOBE}&\textbf{SCAPE}&\textbf{SCAPE}&\textbf{SCAPE}&\textbf{} \\
  \multicolumn{1}{c|}{\textbf{}} &\textbf{to}&\textbf{to}&\textbf{to}&\textbf{to}&\textbf{to}&\textbf{to}&\textbf{to}&\textbf{to}&\textbf{to}&\textbf{to}&\textbf{to}&\textbf{to}&\textbf{Avg.} \\
  \multicolumn{1}{c|}{} &\textbf{ADOBE}&\textbf{MIT}&\textbf{SCAPE}&\textbf{ADOBE}&\textbf{FAUST}&\textbf{SCAPE}&\textbf{FAUST}&\textbf{MIT}&\textbf{SCAPE}&\textbf{ADOBE}&\textbf{FAUST}&\textbf{MIT}& \textbf{} \\
\hline
\hline
DefRec & $79.7~\pm~0.3$ & $61.8~\pm~0.1$ & $\textbf{67.4}~\pm~\textbf{1.0}$ & $67.1~\pm~1.0$& $40.1~\pm~1.4$& $72.6~\pm~0.5$& $42.5~\pm~0.3$& $28.9~\pm~1.5$& $32.2~\pm~1.2$& $66.4~\pm~0.9$& $\textbf{72.2}~\pm~\textbf{1.2}$& $66.2~\pm~0.9$& $58.1~\pm~0.9$\\
DefRec~+~PCM & $78.8~\pm~0.2$ & $60.9~\pm~0.8$ & $63.6~\pm~0.1$ & $48.1~\pm~0.4$& $48.6~\pm~2.4$& $70.1~\pm~0.8$& $46.9~\pm~1.0$& $33.2~\pm~0.3$& $37.6~\pm~0.1$& $66.3~\pm~1.7$& $66.5~\pm~1.0$& $62.6~\pm~0.2$& $56.9~\pm~0.7$\\
\hline
SEN+DefRec & $\textbf{81.9}~\pm~\textbf{0.1}$ & $62.8~\pm~0.3$ & $67.1~\pm~0.4$ & $63.2~\pm~1.5$ & $57.1~\pm~0.8$ & $75.5~\pm~0.3$ & $48.6~\pm~0.8$ & $35.2~\pm~0.9$ & $37.1~\pm~0.5$ & $\textbf{74.1}~\pm~\textbf{0.2}$ & $67.9~\pm~0.8$ & $71.8~\pm~0.1$ & $61.9~\pm~0.6$ \\
SEN+DefRec+PM & $77.1~\pm~0.5$ & $63.0~\pm~0.3$ & $64.5~\pm~0.4$ & $\textbf{79.1}~\pm~\textbf{0.7}$ & $56.6~\pm~1.4$ & $76.1~\pm~0.4$ & $50.8~\pm~1.2$ & $42.8~\pm~0.3$ & $42.2~\pm~0.4$ & $65.9~\pm~0.3$ & $60.8~\pm~0.8$ & $69.2~\pm~0.1$ & $62.3~\pm~0.6$ \\
SEN & $79.7~\pm~0.3$ & $\textbf{63.9}~\pm~\textbf{0.4}$ & $65.1~\pm~0.6$ & $64.7~\pm~1.7$ & $\textbf{58.7}~\pm~\textbf{0.1}$ & $\textbf{76.2}~\pm~\textbf{0.3}$ & $40.5~\pm~0.1$ & $32.0~\pm~0.6$ & $31.9~\pm~0.6$ & $64.9~\pm~0.7$ & $61.0~\pm~1.1$ & $\textbf{72.2}~\pm~\textbf{0.1}$ & $59.2~\pm~0.5$ \\
SEN~+~PM & $75.9~\pm~0.9$ & $63.1~\pm~0.4$ & $62.8~\pm~0.3$ & $78.7~\pm~0.5$ & $56.4~\pm~1.0$ & $75.7~\pm~0.1$ & $\textbf{51.8}~\pm~\textbf{0.8}$ & $\textbf{43.1}~\pm~\textbf{0.4}$ & $\textbf{43.2}~\pm~\textbf{1.0}$ & $64.7~\pm~0.3$ & $67.1~\pm~0.1$ & $69.5~\pm~0.1$ & $\textbf{62.7}~\pm~\textbf{0.5}$ \\
\hline
\end{tabular}}
\end{table*}

\subsection{Ablation study}
We conduct ablation studies on our SEN framework with or without the Point Mixup (PM) models and the proposed consistency loss. The experiment in the DefRec proposed that the different kinds of deformations on 3D point clouds with or without the Point Cloud Mixup (PCM) can achieve the optimal results and also can capture the semantic structures of objects on an unlabeled dataset. We adopt the deformations of shapes proposed by the DefRec to join our proposed SEN for validating their effectiveness on classification and segmentation tasks. Furthermore, we visualize the qualitative analysis of reconstruction results on SEN with/without DefRec and PM methods.

{\bf SEN with additional models on classification.} To verify that SEN is a reasonably good solution for domain adaptation, we compare SEN against several different models on PointDA-10 dataset. Table \ref{tab:table_ca} summarizes the comparison on classification. We can observe that the performance of the SEN is better than the DefRec with or without (w/o) any models. We use one of the best deformations on 3D point clouds for comparison, which is the volume-based deformations with equally sized voxels. Specifically, our SEN with DefRec S/T and PM obtain an improvement over the current state-of-the-art methods by at least $1\%$ on average. And this combination model achieves the best performance on cross-domain scenarios, except for the adaptation from ShapeNet to ModelNet. It is also worth noting that our SEN could achieve competitive results in most scenarios without deformation on 3D point clouds (DefRec).

{\bf SEN with additional models on segmentation.} We then evaluated our method on the PointSegDA dataset with the different models for 3D point cloud segmentation. As shown in Table \ref{tab:table_sa}, our SEN shows its superiority on the domain adaptation segmentation tasks. Compared with the original DefRec w/o PCM, similar improvements in most scenarios can also be observed by adopting SEN w/o different kinds of models on segmentation tasks. Similar to what was observed in the previous experiments, it indicates that: 1) Adding auxiliary models on SEN is an effective strategy to boost the performance of the network for UDA on 3D point clouds. 2) The competitive results have validated the applicability of SEN, which doesn't rely on the deformations for 3D point cloud domain adaptation tasks.

\begin{figure}
\center
\includegraphics[width=0.9\linewidth]{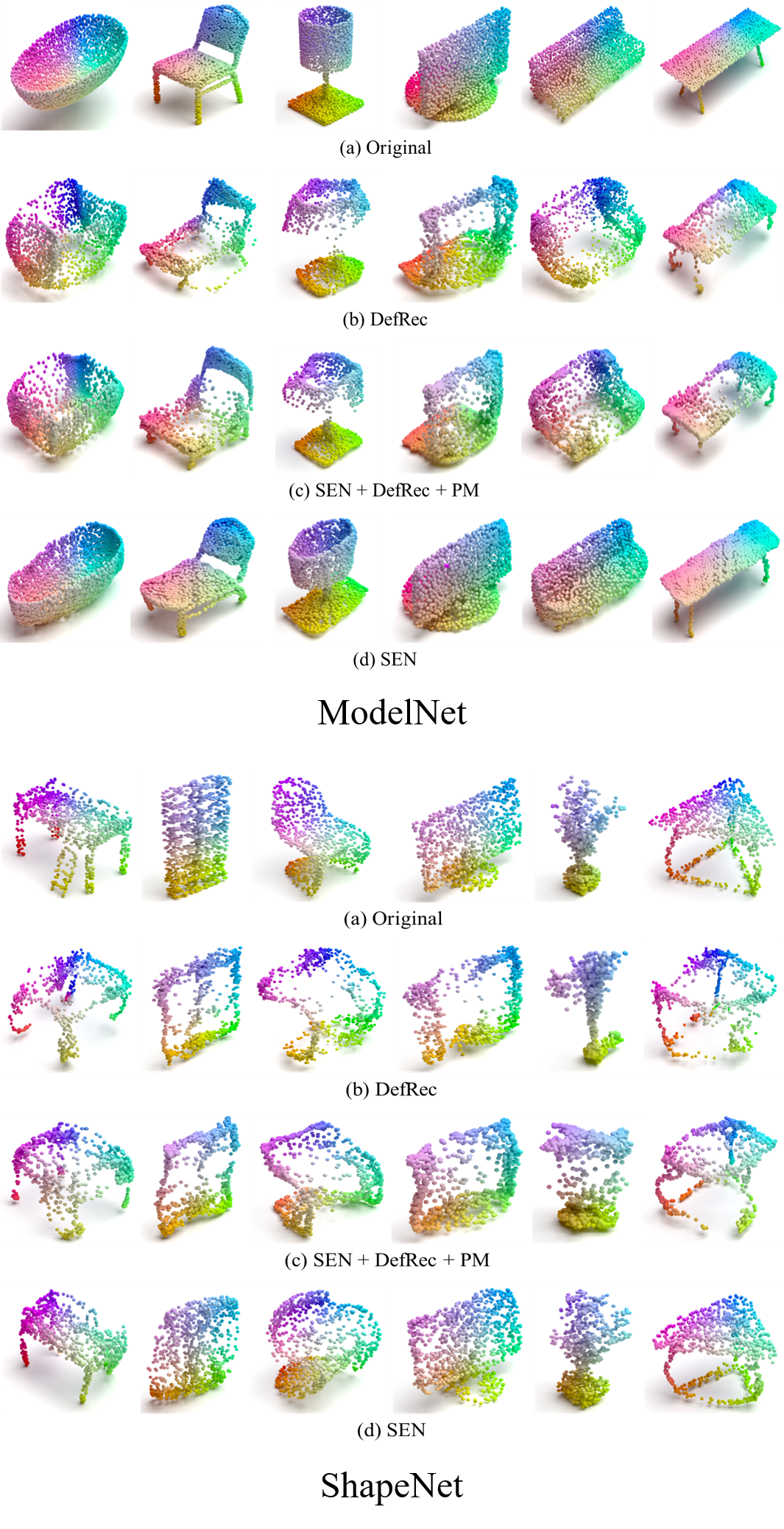}
\caption{Visualization of the reconstruction results by our SEN with and without (w/o) DefRec and PM module are cross-domain between the ModelNet-10 and ShapeNet-10 subsets. The ModelNet-10 and ShapeNet-10 present the reconstruction visualization on the corresponding target domain. The reconstruction samples are sampled from the different classes and obtained by the SEN w/o DefRec and PM method, and the DefRec method. The original represents the original point cloud samples.}
\label{fig:sg_v}
\end{figure}

{\bf Qualitative analysis of reconstruction results.} To further illustrate the effectiveness of our SEN, we compare the reconstruction results with two different methods (like DefRec and SEN+DefRec+PM) in Figure \ref{fig:sg_v}. We show the results of the original and DefRec methods on ModelNet-10 and ShapNet-10 datasets in the (a) and (b) row, respectively. We also illustrate the reconstruction performance of our SEN with and without the DefRec and PM module in the (c) and (d) row, respectively. The visualization of reconstruction results shows that our SEN could capture a better structural representation in the form of 3D cloud points. Such as, the SEN (d) clearly exhibits more fine structures and details in ModelNet-10, because SEN captures the more discriminative semantic features. Particularly, the SEN doesn't need to rely on any other deformation approaches. This experiment indicates that: 1) the collaborative self-ensembling is able to learn the discriminative feature representations on 3D domain adaptation tasks. 2) the SEN utilizes the teacher network to guide the student network learning in the target domain so that can improve the quality of point clouds reconstruction, which consequently boosts the student net's classification performance.
% the temporal consistency to construct the teacher net to learn the discriminative feature representations and ensure the quality of point clouds reconstruction.
% %by guidance learning from a teacher network to enhance the abilities of student network

\section{Conclusion}
In this paper, we proposed a novel self-ensembling network (SEN) framework for 3D point cloud domain adaptation tasks. It mainly leveraged the idea of Mean Teacher to construct the SEN network to transfer knowledge gained from a labeled source dataset to a distinct unlabeled target dataset. Specifically, we introduce a soft classification loss and a consistency loss, which aim to achieve the consistency generalization between the source and the target domain and improve the accurate reconstruction on the target domains. Our SEN can train the student network collaboratively with supervised learning and self-supervised learning and utilizes the teacher network to conduct the temporal consistency to learn the useful feature representations and ensure the quality of point clouds reconstruction. Extensive experiments on two challenging 3D domain adaptation datasets demonstrated the effectiveness of our SEN framework on classification and segmentation tasks. Our SEN framework outer performs state-of-the-art performance for 3D point cloud domain adaptation tasks.

{\small
\bibliographystyle{ieee}
\bibliography{ref}
}

\end{document}